\documentclass{article}
\usepackage{spconf,amsmath,epsfig}
\usepackage{url}
\usepackage{xcolor}
\usepackage{multirow} 
\usepackage{array} 
\newcolumntype{C}[1]{>{\centering\arraybackslash}p{#1}}
\title{Mapping Earth Mounds from Space}
\twoauthors
  {B.\ Uzun, S.\ Pande, G.\ Cachin-Bernard, M.T.\ Pham, S.\ Lef\`evre}
	{Universit\'e Bretagne Sud, UMR IRISA 6074\\
	Vannes, France}
  {R.\ Blatrix, D.\ McKey\thanks{This research was funded by the TOSCA committee (Terre Solide, Océan, Surfaces Continentales, Atmosphère) of the CNES (French National Center for Space Research) through the grant named FLOODSCAPE, and by the Institut Ecologie et Environnement (INEE) / CNRS (Projets Exploratoires Pluridisciplinaires TOHMIS) through the grant named INOCULT.}}
	{CEFE, Univ Montpellier, CNRS, EPHE, IRD\\
	Montpellier, France}
\begin{document}
%
\maketitle
\begin{abstract}
Regular patterns of vegetation are considered widespread landscapes, although their global extent has never been estimated. Among them, spotted landscapes are of particular interest in the context of climate change. Indeed, regularly spaced vegetation spots in semi-arid shrublands result from extreme resource depletion and prefigure catastrophic shift of the ecosystem to a homogeneous desert, while termite mounds also producing spotted landscapes were shown to increase robustness to climate change. Yet, their identification at large scale calls for automatic methods, for instance using the popular deep learning framework, able to cope with a vast amount of remote sensing data, e.g., optical satellite imagery. In this paper, we tackle this problem and benchmark some state-of-the-art deep networks on several landscapes and geographical areas. Despite the promising results we obtained, we found that more research is needed to be able to map automatically these earth mounds from space.
\end{abstract}
\section{Introduction}
\label{sec:intro}

Regular patterns of vegetation are considered widespread landscapes, although their global extent has never been estimated. They show a great diversity of shapes, including labyrinths, strips, gaps and spots, the formation of which can be explained by scale-dependent feedbacks between organisms and limiting resources. Water as a limiting resource has received particular attention in cases studied to date \cite{rietkerk2002self,rietkerk2004self,rietkerk2008regular}. Territorial competition between colonies of social-insect ecosystem engineers (termites in particular) has also been shown to be another mechanism underlying the formation of vegetation spots and gaps \cite{pringle2017spatial}, with a distinct theoretical framework \cite{tarnita2017theoretical}. Both mechanisms are not mutually exclusive.

Among the diversity of shapes observed in patterned landscapes, spotted landscapes are the most frequent and widespread, and are also the best studied. Regularly spaced vegetation spots in semi-arid shrublands result from extreme resource depletion and prefigure catastrophic shift of the ecosystem to a homogeneous desert \cite{rietkerk2004self,scheffer2009early}. In addition, termite mounds producing spotted landscapes were shown to increase robustness to climate change \cite{bonachela2015termite}. Thus, spotted landscapes are of particular interest not only to foresee potential consequences of climate change, but also to buffer the effects of climate change by delaying catastrophic shift. 

Although spotted landscapes have been studied mostly in dry environments, they are also widespread in seasonally flooded savannas. Both environments have in common a strong constraint on ecosystem engineers: water is limited in dry environments, well-aerated soil is limited in flooded environments, and low soil fertility often characterizes both. Similar processes may thus drive landscape patterns in both environments and the aforementioned theoretical frameworks may similarly apply. Spotted landscapes are often associated with microtopographic heterogeneity: spots are earthmounds, typically ranging from 1 to 30 m in diameter and 0.2 to 4 m in height. Although present in dry environments, earthmounds are of particular importance in flooded environments, where even slight differences in elevation lead to highly contrasted levels of the constraining resource (well-aerated soils). Thus, understanding the drivers of microtopographic heterogeneity should help to account for the dynamics of most spotted landscapes, in both semi-arid and seasonally flooded landscapes. Various origins have been proposed for the formation of earthmounds, e.g., differential erosion, eolian deposition of sediment, burrowing activity of mammals, nest construction by termites, past or current human agency (raised-field agriculture), leading to a variety of regional names of these landscapes in the literature, e.g., Heuweltjies, Nabkha, Mima mounds, Murundus, termite savannas, raised fields \cite{cramer2014mima}. In many cases, several mechanisms may have been in play successively, or may be in play simultaneously, making it difficult to identify the original cause.

Whatever their origin, earthmound landscapes share common properties that enhance ecosystem functioning and thus make them of particular interest for researchers in ecology in the context of current global changes. The first property is that each mound works as a fertility island \cite{souza2020termite,pringle2010spatial,kunz2012effects,renard2013ancient,marimon2015ecology}. Various mechanisms, not mutually exclusive, may account for fertility islands. Feedback loops between short-range facilitation and long-range competition for resources between individual plants result in resource concentration under clumps of vegetation \cite{schlesinger1990biological}. Central-place foragers such as social insects bring food to their nest, leading to the concentration of organic matter in or close to the nest. As a consequence, large termite nests in termite savannas, for instance, have higher contents of soil nutrients than the surrounding savanna soil, with cascading effects on biodiversity and productivity \cite{pringle2010spatial,sileshi2010termite}. Raised-field agriculture is a current practice in some parts of Africa, consisting in piling up topsoil and organic matter to locally increase nutrient availability to crop, the consequence of which is increased yield \cite{comptour2018wetland,rodrigues2020congo}. Raised-field agriculture has been extensively practiced in South America by pre-Columbian civilizations \cite{denevan2001cultivated}.
Although this practice has been abandoned for several hundred years in the Neotropics, vestiges of raised fields are still visible because the topographic heterogeneity initiated by humans has been beneficial to ecosystem engineers (termites, ants, earthworms, plants) that have maintained resource concentration (and the mounds themselves) through feedbacks ever since the fields were abandoned \cite{mckey2010pre,renard2013ancient}. Earthmounds, as fertility islands, increase biodiversity and productivity of the ecosystem.  The second property common to earthmound landscapes is the regular spacing between earthmounds \cite{pringle2010spatial,renard2012origin,cramer2015distribution,de2018fine}. The effect of regular spacing on ecosystem functioning has been little investigated, but it was shown that regular spacing of termite mounds increased ecosystem productivity compared to random spacing because it optimizes mound density \cite{pringle2010spatial}.

The increasing ease of access to satellite images at resolutions high enough to detect meter-scale landscape features opens unprecedented opportunities to study regular patterns of vegetation, and in particular spatial patterns of earthmound landscapes, at the global scale. However, images have to be visually checked and mounds individually labeled for applying point pattern analysis. Thus, characterizing spatial patterns of earthmound landscapes at a global scale requires labor-intensive manual processing. Automatic image processing is thus needed in ecology to reduce manual labor and produce large sets of reliable data, in particular from remote sensing sources.
To do so, deep learning has become the reference methodology for identifying patterns in images (including ecological patterns \cite{brodrick2019}), following the general trend in computer vision and various application fields. Yet, applying deep learning in environmental remote sensing raises some challenges \cite{yuan2020}, especially the need for large training sets, that can be countered with various strategies \cite{safonava2023,bjorck2021}.

The problem of extracting earth mounds from remote sensing using deep neural networks has been already tackled by a few studies. In \cite{brodrick2019}, termite mounds are identified with a simple CNN as circular shapes clearly visible in a DEM generated from LiDAR data. In \cite{sales2021}, the problem is expressed as an object detection task solved with RetinaNet using data acquired with a GoPro camera. More closely related to our work, \cite{lind2019,lind2021} process very high resolution satellite imagery through CNNs, but no details have been provided on the model and the experimental settings. The problem of automatic earth mound identification from space using optical imagery and deep learning remains still largely unexplored. 

We aim here to fill this gap through assessing several state-of-the-art deep networks for this purpose, considering several kinds of landscapes and geographical areas. We show that the problem is complex and requires further research.

\section{Material and methods}

\subsection{Study sites}

\begin{figure}
\includegraphics[width=\columnwidth]{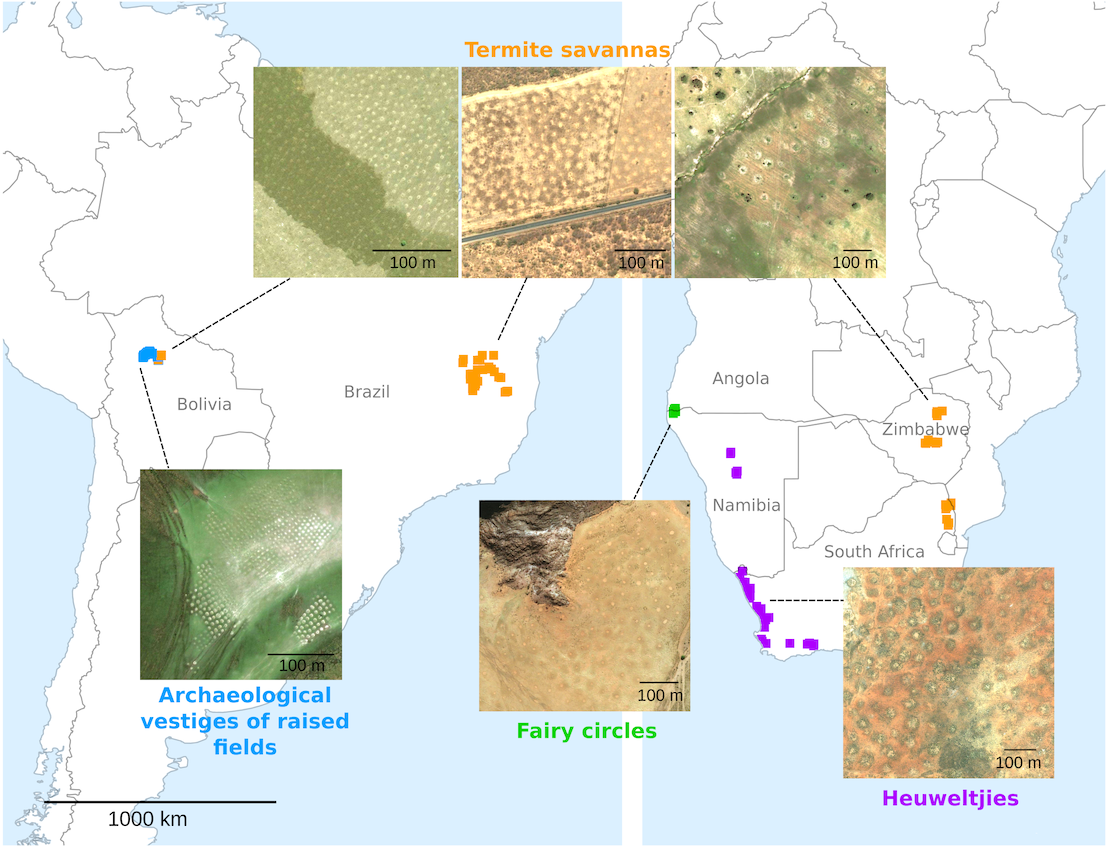}
\caption{Sites in South America and Africa from which satellite images were extracted, with illustration of the four categories of spotted landscapes. Images are samples from Maxar WorldView-3 extracted from Microsoft Bing Maps.}
\label{fig1}
\end{figure}

The spotted landscapes we studied were located in South America and Africa and were of four commonly recognized categories (Fig.~\ref{fig1}).

\subsubsection{Termite savannas}

\begin{figure}
\includegraphics[width=\columnwidth]{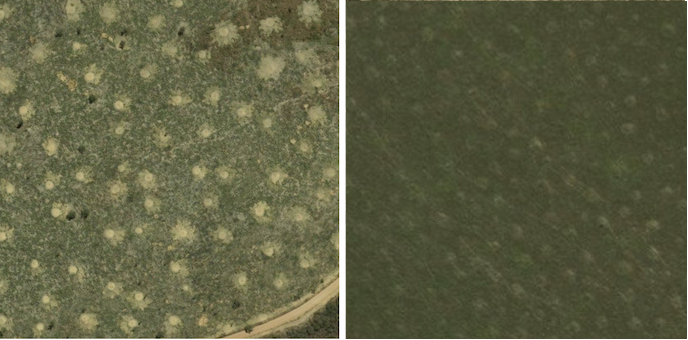}
\caption{Examples of satellite images of a termite savanna in Brazil illustrating the two levels of detection difficulty (easy on the left image, difficult on the right image). Same scale for both images. Satellite samples have been taken from Microsoft Bing Maps (Maxar WorldView-3 data).}
\label{fig2}
\end{figure}

Termite savannas feature large termite mounds (about 20 m in diameter and 2 m high) that are regularly spaced and support distinct vegetation compared to the surrounding area. Each mound houses a single termite colony, with the nearest neighbor over 50 m away. We selected sites in Bolivia (28), Brazil (2), South Africa (8), and Zimbabwe (11). Another type of termite savanna, recently discovered in Brazil, spans an area nearly the size of Great Britain \cite{funch2015termite,martin2018vast}. Here, mounds are closer together and result from millennia of soil excavation from underground tunnels. Unlike typical termite nests, these mounds lack a network of tunnels and chambers. We selected 29 sites from this unique landscape. Due to their distinct characteristics, we classified them separately as ``supercolony.'' Fig.~\ref{fig2}  illustrates the size variations of termite mounds across Brazil, leading to various levels of difficulty for the detection task. Our study also highlights the significant differences in termite mound structures across continents.


\subsubsection{Heuweltjies}

Although the origin of Heuweltjies is still debated (eolian deposition of sediments or termite mounds), this type of landscape shares similarities with termite savannas (size and spacing of the mounds in particular) \cite{cramer2015distribution}. This landscape type is only known from south-western Africa. We selected sites in South Africa (32) and Namibia (5).

\subsubsection{Fairy circles}

This landscape type looks like a negative of termite savannas: spots are depressions of bare soil surrounded by vegetation. As for Heuweltjies, their origin is debated. The two best supported hypotheses suggest that fairy circles result either from vegetation-water feedbacks \cite{getzin2016discovery} or from the activity of subterranean termite nests \cite{juergens2013biological}. They are known from south-western Africa and Australia. We selected sites in Namibia (3) and Angola (2).

\subsubsection{Archaeological vestiges of raised fields}

Raised fields constructed by pre-Columbian civilizations in South America took various forms \cite{denevan2001cultivated}, one of which is circular mounds. The vestiges of circular raised fields have a spatial organization strikingly different from that of termite savannas: mounds are less than a few meters in diameter and usually (today) less than 0.5 m high; the nearest neighbor is within a few meters and in addition to showing regular spacing; they are often oriented in square grids \cite{renard2012origin}. We selected 93 sites in Bolivia following the map of Rodrigues et al.~\cite{rodrigues2018design}. 

For each study site, we extracted a square area of c.\ 0.58~km2 using Microsoft Bing Maps at the highest resolution (corresponding to Maxar WorldView-3 imagery). Following visual assessment of the study sites, we considered only a subset of the selected sites, ignoring images for which the annotation was questionable.

\subsection{Automatic image analysis workflow}

\subsubsection{Data preparation}
We split our data in two parts following the nature of the mounds, i.e.\ due to human factors or not. The first dataset, called Raised fields dataset in the sequel, includes 67 images with a spatial extent of 2560$\times $2560 pixels. The mounds are small and densely packed, making individual annotation extremely time-consuming. Therefore, we resorted to annotating the group of mounds as a single entity/instance.

The second dataset, called the Termites dataset, contains 76 images featuring four classes: termite savannas (containing only single termite colonies),  supercolonies (a more complex kind of termite savannas), fairy circles, and Heuweltjies. Conversely to the first dataset, we annotated each mound individually, given the sparse nature of the mounds. 

To prevent memory issues, we resized the images in the Raised fields dataset to 1024$\times$1024 pixels. For the Termites dataset, we split the images into 640$\times$640 tiles. The annotations of the mounds in all the images were carried out using \emph{Make-Sense.ai} annotation tool \cite{makesense}.


\subsubsection{Deep learning architectures}
We aim to assess the performance of well-established deep architectures for object detection in our specific context. We selected one of the most recent models from the YOLO familly \cite{redmon2016you}, namely YOLOv8 \cite{Jocher_Ultralytics_YOLO_2023} (with the original model, medium size), as well as Faster R-CNN \cite{ren2016faster}, using two different backbones: ResNet50 \cite{he2015deep} and MobileNet \cite{howard2017mobilenets}. 

The choice of the two backbones was done in order to assess the efficiency of the models in a practical scenario, where both accuracy and deployment time are significant. Let us recall that YOLOv8 comes under the category of one-stage detector models, where the images are processed in a single pass, and the class probability prediction and bounding box estimation happened simultaneously. Such a strategy leads to faster deployment, but the model struggles with highly accurate detection, especially for small objects. On the other hand, we are considering Faster R-CNN, a two-stage detector. In the first stage, the region proposals are generated, while in the second stage, these proposals are fine-tuned and classified. This makes the model highly accurate and more robust for small object detection, but it comes at a higher computational cost and thus a lower processing speed.

\subsubsection{Training protocols and evaluation}

We rely on a standard evaluation protocol, and split our dataset into a training set (80\%) and a testing set (20\%). All the models are trained using AdamW \cite{loshchilov2017decoupled} optimizer, with a learning rate of 0.001. During training, the batch size has been set to 16. 
The performance of the models is evaluated using average precision and mean average precision at 0.5 IoU threshold (mAP50).


\section{Results and Discussion}

Tables \ref{tab:rf_perf} and \ref{tab:term_perf} respectively showcase the detection performance on the Raised fields and Termites datasets. The highest reported values are boldened. Also, for both datasets, the performance of YOLOv8 model is reported with and without augmentation (as YOLOv8 was used as an off-the-shelf method and the code supported automatic data augmentation). 

It is observed that in case of the raised fields dataset, the best performance is achieved with Faster R-CNN with the ResNet50 backbone. While older, this method surpasses the YOLOv8 model, even with augmentation. The reason could be that the mounds in the images are relatively small (within a bounding box), and YOLOv8 does not particularly specialises in detecting small objects. On the contrary, YOLOv8 surpasses Faster R-CNN on the Termites dataset, probably due to large size of mounds and lighter architecture. 

\begin{figure}
\includegraphics[width=\columnwidth]{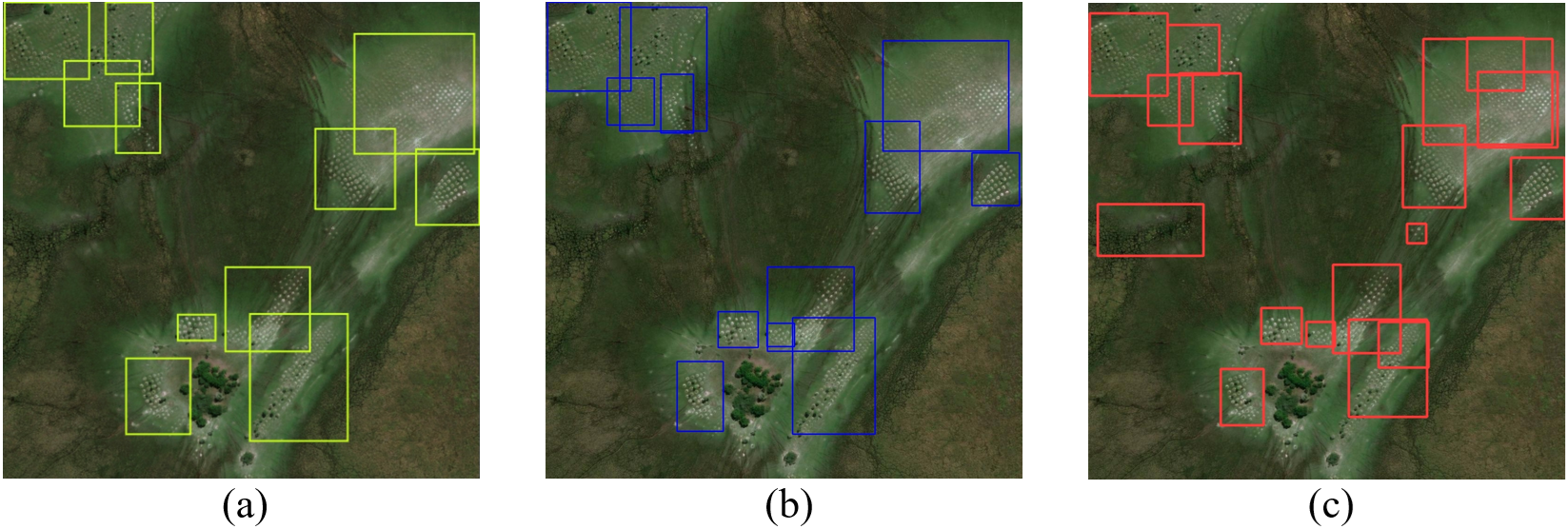}
\caption{Visual illustration of results obtained on the Raised fields dataset: (a) Groundtruth, and detection maps generated with (b) Faster R-CNN and (c) YOLOv8.}
\label{fig:od_rf}
\end{figure}

\begin{figure}
\includegraphics[width=\columnwidth]{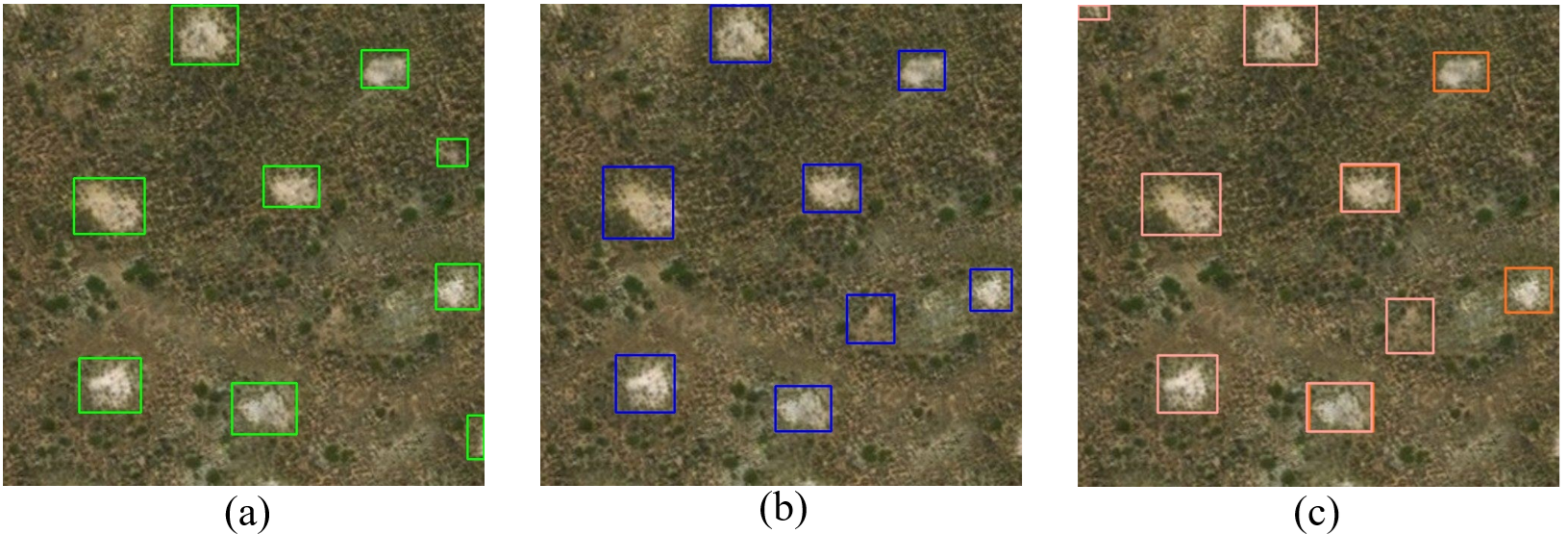}
\caption{Visual illustration of results obtained on the Termites dataset (Heuweltjies class): (a) Groundtruth, and detection maps generated with (b) Faster R-CNN and (c) YOLOv8.}
\label{fig:od_t}
\end{figure}

We also report some qualitative results on the two datasets in Figures \ref{fig:od_rf} and \ref{fig:od_t}. We can see that for the Raised fields dataset, YOLOv8 tended to overpredict the number of mound areas w.r.t.\ Faster R-CNN, thus generating a lot of false positives. This could be due to difficulty in learning the smaller size of mounds and confusing the `non-mound' area with those of mounds. For the Termites dataset, we see that both models robustly localise the mounds for the Heuweltjies class, though some smaller mound still go unnoticed. Deep models dedicated to small objects \cite{rs12193152,rs12152501} should be considered for such specific settings.

Moreover, we also conducted an ablation study to assess the performance when the number of training samples are decreased. To do so, we considered YOLOv8 model since it is a more efficient model. As observed in Tab.~\ref{tab:abl_ns}, the accuracy decreases for some classes when the number of samples is reduced from 60 to 50. The degradation of the results is more noticeable when the number of training samples is further reduced to 40. But even with only 40 samples, YOLOv8 is competing Faster R-CNN (see Tab.~\ref{tab:term_perf}).

\begin{table}[t!]
    \centering
    {\scriptsize
    \caption{\label{tab:rf_perf} Quantitative evaluation on the Raised fields dataset.}
    \begin{tabular}
    {c|c|c|c|c}
        & Faster R-CNN & Faster R-CNN & YOLOv8 & YOLOv8\\
        & (ResNet) & (MobileNet) & (No Aug) & (with Aug)\\ \hline
        Overall mAP & \textbf{51.4} & 44.3 & 28.8 &42.8\\ 
    \end{tabular}}
\end{table}

\begin{table}[t!]
    \centering
    {\scriptsize
    \caption{\label{tab:term_perf} Quantitative evaluation on the Termites dataset.}
    \begin{tabular}
    {c|c|c|c|c}
        \multirow[b]{2}{*}{Classes} & Faster R-CNN & Faster R-CNN & YOLOv8 & YOLOv8\\
        & (ResNet) & (MobileNet) & (No Aug) & (with Aug)\\ \hline
        Fairy Circles & 16.7 & 29.3 & 78.8 & \textbf{85.7}\\ 
        Heuweltjies & 54.8 & \textbf{55.1} & 37.4 & 37.6\\ 
        Supercolony& 12.5 & 25.7 & 23.5 & \textbf{58.9}\\ 
        Savannas& 43.3 & 33.0 & 24.6 & \textbf{50.4}\\ \hline
        Overall mAP & 31.8 & 35.8 & 42.9 &\textbf{58.2}\\ 
    \end{tabular}}
\end{table}

\begin{table}[t!]
    \centering
    {\scriptsize
    \caption{\label{tab:abl_ns}Ablation study to assess the impact of the number of samples on the performance of YOLOv8 with augmentation.}
    \begin{tabular}{c|c|c|c}
        Classes & 40 Samples& 50 Samples& 60 Samples\\ \hline
        Fairy Circles & 77.2 & 88.7 & 85.7 \\ 
        Heuweltjies & 20.5 & 36.9 & 37.6 \\ 
        Supercolony& 10.9 & 20.8 & 58.9 \\ 
        Savannas & 31.2 & 50.2 & 50.4 \\ \hline
        Overall mAP & 34.9 & 49.1 & 58.2 \\ 
    \end{tabular}}
\end{table}




\section{Conclusion}
In this paper, we aim at detecting mounds from high-resolution remote sensing satellite imagery. We distinguish between artificial farming mounds (a.k.a.\ raised fields) and natural ones. Using two popular models, namely Faster R-CNN and YOLOv8, we found that YOLOv8 performed better for natural mounds, while Faster R-CNN was superior for the Raised fields. Despite their effectiveness, geographical, spectral, and spatial variability still poses challenges for perfect localization. Future research will explore advanced deep learning techniques and domain-invariant learning to improve model transferability across different geographical locations.

\clearpage
\bibliographystyle{IEEEbib}
{\footnotesize\bibliography{refs}}

\begin{thebibliography}{10}

\bibitem{rietkerk2002self}
Max Rietkerk, Maarten~C Boerlijst, Frank van Langevelde, Reinier HilleRisLambers, Johan~van de~Koppel, Lalit Kumar, Herbert~HT Prins, and Andr{\'e}~M de~Roos,
\newblock ``Self-organization of vegetation in arid ecosystems,''
\newblock {\em The American Naturalist}, vol. 160, no. 4, pp. 524--530, 2002.

\bibitem{rietkerk2004self}
Max Rietkerk, Stefan~C Dekker, Peter~C De~Ruiter, and Johan van~de Koppel,
\newblock ``Self-organized patchiness and catastrophic shifts in ecosystems,''
\newblock {\em Science}, vol. 305, no. 5692, pp. 1926--1929, 2004.

\bibitem{rietkerk2008regular}
Max Rietkerk and Johan Van~de Koppel,
\newblock ``Regular pattern formation in real ecosystems,''
\newblock {\em Trends in Ecology \& Evolution}, vol. 23, no. 3, pp. 169--175, 2008.

\bibitem{pringle2017spatial}
Robert~M Pringle and Corina~E Tarnita,
\newblock ``Spatial self-organization of ecosystems: integrating multiple mechanisms of regular-pattern formation,''
\newblock {\em Annual review of Entomology}, vol. 62, pp. 359--377, 2017.

\bibitem{tarnita2017theoretical}
Corina~E Tarnita, Juan~A Bonachela, Efrat Sheffer, Jennifer~A Guyton, Tyler~C Coverdale, Ryan~A Long, and Robert~M Pringle,
\newblock ``A theoretical foundation for multi-scale regular vegetation patterns,''
\newblock {\em Nature}, vol. 541, no. 7637, pp. 398--401, 2017.

\bibitem{scheffer2009early}
Marten Scheffer, Jordi Bascompte, William~A Brock, Victor Brovkin, Stephen~R Carpenter, Vasilis Dakos, Hermann Held, Egbert~H Van~Nes, Max Rietkerk, and George Sugihara,
\newblock ``Early-warning signals for critical transitions,''
\newblock {\em Nature}, vol. 461, no. 7260, pp. 53--59, 2009.

\bibitem{bonachela2015termite}
Juan~A Bonachela, Robert~M Pringle, Efrat Sheffer, Tyler~C Coverdale, Jennifer~A Guyton, Kelly~K Caylor, Simon~A Levin, and Corina~E Tarnita,
\newblock ``Termite mounds can increase the robustness of dryland ecosystems to climatic change,''
\newblock {\em Science}, vol. 347, no. 6222, pp. 651--655, 2015.

\bibitem{cramer2014mima}
Michael~D Cramer and Nichole~N Barger,
\newblock ``Are mima-like mounds the consequence of long-term stability of vegetation spatial patterning?,''
\newblock {\em Palaeogeography, Palaeoclimatology, Palaeoecology}, vol. 409, pp. 72--83, 2014.

\bibitem{souza2020termite}
Henrique Jesus~de Souza, Jacques Hubert~Charles Delabie, and George~Andrade Sodr{\'e},
\newblock ``Termite participation in the soil-forming processes of'murundus' structures in the semi-arid region of brazil,''
\newblock {\em Revista Brasileira de Ci{\^e}ncia do Solo}, vol. 44, 2020.

\bibitem{pringle2010spatial}
Robert~M Pringle, Daniel~F Doak, Alison~K Brody, Rudy Jocqu{\'e}, and Todd~M Palmer,
\newblock ``Spatial pattern enhances ecosystem functioning in an african savanna,''
\newblock {\em PLoS biology}, vol. 8, no. 5, pp. e1000377, 2010.

\bibitem{kunz2012effects}
NS~Kunz, MT~Hoffman, and B~Weber,
\newblock ``Effects of heuweltjies and utilization on vegetation patterns in the succulent karoo, south africa,''
\newblock {\em Journal of Arid Environments}, vol. 87, pp. 198--205, 2012.

\bibitem{renard2013ancient}
Delphine Renard, Jago~Jonathan Birk, Anne Zangerl{\'e}, Patrick Lavelle, Bruno Glaser, Rumsa{\"\i}s Blatrix, and Doyle McKey,
\newblock ``Ancient human agricultural practices can promote activities of contemporary non-human soil ecosystem engineers: A case study in coastal savannas of french guiana,''
\newblock {\em Soil Biology and Biochemistry}, vol. 62, pp. 46--56, 2013.

\bibitem{marimon2015ecology}
Beatriz~Schwantes Marimon, Guarino~R Colli, Ben~Hur Marimon-Junior, Henrique~A Mews, Pedro~V Eisenlohr, Ted~R Feldpausch, and Oliver~L Phillips,
\newblock ``Ecology of floodplain campos de murundus savanna in southern amazonia,''
\newblock {\em International Journal of Plant Sciences}, vol. 176, no. 7, pp. 670--681, 2015.

\bibitem{schlesinger1990biological}
William~H Schlesinger, James~F Reynolds, Gary~L Cunningham, Laura~F Huenneke, Wesley~M Jarrell, Ross~A Virginia, and Walter~G Whitford,
\newblock ``Biological feedbacks in global desertification,''
\newblock {\em Science}, vol. 247, no. 4946, pp. 1043--1048, 1990.

\bibitem{sileshi2010termite}
Gudeta~W Sileshi, MA~Arshad, Souleymane Konat{\'e}, and Philip~OY Nkunika,
\newblock ``Termite-induced heterogeneity in african savanna vegetation: mechanisms and patterns,''
\newblock {\em Journal of Vegetation Science}, vol. 21, no. 5, pp. 923--937, 2010.

\bibitem{comptour2018wetland}
Marion Comptour, Sophie Caillon, Leonor Rodrigues, and Doyle McKey,
\newblock ``Wetland raised-field agriculture and its contribution to sustainability: Ethnoecology of a present-day african system and questions about pre-columbian systems in the american tropics,''
\newblock {\em Sustainability}, vol. 10, no. 9, pp. 3120, 2018.

\bibitem{rodrigues2020congo}
Leonor Rodrigues, Tobias Sprafke, Carine Bokatola~Moyikola, Bernard~G Barth{\`e}s, Isabelle Bertrand, Marion Comptour, St{\'e}phen Rostain, Joseph Yoka, and Doyle McKey,
\newblock ``A congo basin ethnographic analogue of pre-columbian amazonian raised fields shows the ephemeral legacy of organic matter management,''
\newblock {\em Scientific Reports}, vol. 10, no. 1, pp. 10851, 2020.

\bibitem{denevan2001cultivated}
William~M. Denevan,
\newblock {\em Cultivated landscapes of Native Amazonia and the Andes},
\newblock Oxford University Press, 2001.

\bibitem{mckey2010pre}
Doyle McKey, St{\'e}phen Rostain, Jos{\'e} Iriarte, Bruno Glaser, Jago~Jonathan Birk, Irene Holst, and Delphine Renard,
\newblock ``Pre-columbian agricultural landscapes, ecosystem engineers, and self-organized patchiness in amazonia,''
\newblock {\em Proceedings of the National Academy of Sciences}, vol. 107, no. 17, pp. 7823--7828, 2010.

\bibitem{renard2012origin}
Delphine Renard, Jago~Jonathan Birk, Bruno Glaser, Jos{\'e} Iriarte, Gilles Grisard, Johannes Karl, and Doyle McKey,
\newblock ``Origin of mound-field landscapes: a multi-proxy approach combining contemporary vegetation, carbon stable isotopes and phytoliths,''
\newblock {\em Plant and Soil}, vol. 351, pp. 337--353, 2012.

\bibitem{cramer2015distribution}
Michael~D Cramer and Jeremy~J Midgley,
\newblock ``The distribution and spatial patterning of mima-like mounds in south africa suggests genesis through vegetation induced aeolian sediment deposition,''
\newblock {\em Journal of Arid Environments}, vol. 119, pp. 16--26, 2015.

\bibitem{de2018fine}
Henrique~J De~Souza and Jacques~HC Delabie,
\newblock ``Fine-scale spatial distribution of murundus structures in the semi-arid region of brazil,''
\newblock {\em Austral Ecology}, vol. 43, no. 3, pp. 268--279, 2018.

\bibitem{brodrick2019}
Philip~G. Brodrick, Andrew~B. Davies, and Gregory~P. Asner,
\newblock ``Uncovering ecological patterns with convolutional neural networks,''
\newblock {\em Trends in Ecology \& Evolution}, vol. 34, no. 8, pp. 734--745, 2019.

\bibitem{yuan2020}
Qiangqiang Yuan, Huanfeng Shen, Tongwen Li, Zhiwei Li, Shuwen Li, Yun Jiang, Hongzhang Xu, Weiwei Tan, Qianqian Yang, Jiwen Wang, Jianhao Gao, and Liangpei Zhang,
\newblock ``Deep learning in environmental remote sensing: Achievements and challenges,''
\newblock {\em Remote Sensing of Environment}, vol. 241, pp. 111716, 2020.

\bibitem{safonava2023}
Anastasiia Safonova, Gohar Ghazaryan, Stefan Stiller, Magdalena Main-Knorn, Claas Nendel, and Masahiro Ryo,
\newblock ``Ten deep learning techniques to address small data problems with remote sensing,''
\newblock {\em International Journal of Applied Earth Observation and Geoinformation}, vol. 125, pp. 103569, 2023.

\bibitem{bjorck2021}
Johan Bjorck, Brendan~H. Rappazzo, Qinru Shi, Carrie Brown-Lima, Jennifer Dean, Angela Fuller, and Carla Gomes,
\newblock ``Accelerating ecological sciences from above: Spatial contrastive learning for remote sensing,''
\newblock {\em Proceedings of the AAAI Conference on Artificial Intelligence}, vol. 35, no. 17, pp. 14711--14720, May 2021.

\bibitem{sales2021}
Juan Sales, José Marcato~Junior, Henrique Siqueira, Mauricio De~Souza, Edson Matsubara, and Wesley~Nunes Gonçalves,
\newblock ``Retinanet deep learning-based approach to detect termite mounds in eucalyptus forests,''
\newblock in {\em IEEE International Geoscience and Remote Sensing Symposium}, 2021, pp. 586--589.

\bibitem{lind2019}
B.~M. {Lind}, M.~{Brandt}, A.~{Kariryaa}, J.~L. {Small}, K.~A. {Melocik}, C.~J. {Tucker}, and N.~P. {Hanan},
\newblock ``{Very High Resolution Satellite Imagery Reveals Many Millions of Termite Mounds Across the West African Sahel},''
\newblock in {\em AGU Fall Meeting Abstracts}, Dec. 2019, vol. 2019, pp. B23F--2611.

\bibitem{lind2021}
Brianna~M. {Lind}, Martin {Brandt}, Ankit {Kariryaa}, Jennifer {Small}, Katherine {Melocik}, Compton {Tucker}, and Niall {Hanan},
\newblock ``{Deep Learning and High Resolution Satellite Imagery Reveal 10s of Millions of Termite Mounds in Senegal, West Africa},''
\newblock in {\em AGU Fall Meeting Abstracts}, Dec. 2021, vol. 2021, pp. B51B--04.

\bibitem{funch2015termite}
Roy~Richard Funch,
\newblock ``Termite mounds as dominant land forms in semiarid northeastern brazil,''
\newblock {\em Journal of Arid Environments}, vol. 122, pp. 27--29, 2015.

\bibitem{martin2018vast}
Stephen~J Martin, Roy~R Funch, Paul~R Hanson, and Eun-Hye Yoo,
\newblock ``A vast 4,000-year-old spatial pattern of termite mounds,''
\newblock {\em Current Biology}, vol. 28, no. 22, pp. R1292--R1293, 2018.

\bibitem{getzin2016discovery}
Stephan Getzin, Hezi Yizhaq, Bronwyn Bell, Todd~E Erickson, Anthony~C Postle, Itzhak Katra, Omer Tzuk, Yuval~R Zelnik, Kerstin Wiegand, Thorsten Wiegand, et~al.,
\newblock ``Discovery of fairy circles in australia supports self-organization theory,''
\newblock {\em Proceedings of the National Academy of Sciences}, vol. 113, no. 13, pp. 3551--3556, 2016.

\bibitem{juergens2013biological}
Norbert Juergens,
\newblock ``The biological underpinnings of namib desert fairy circles,''
\newblock {\em Science}, vol. 339, no. 6127, pp. 1618--1621, 2013.

\bibitem{rodrigues2018design}
Leonor Rodrigues, Umberto Lombardo, and Heinz Veit,
\newblock ``Design of pre-columbian raised fields in the llanos de moxos, bolivian amazon: Differential adaptations to the local environment?,''
\newblock {\em Journal of Archaeological Science: Reports}, vol. 17, pp. 366--378, 2018.

\bibitem{makesense}
Piotr Skalski,
\newblock ``{Make Sense},'' 2019,
\newblock \url{https://github.com/SkalskiP/make-sense/}.

\bibitem{redmon2016you}
Joseph Redmon, Santosh Divvala, Ross Girshick, and Ali Farhadi,
\newblock ``You only look once: Unified, real-time object detection,''
\newblock in {\em Proceedings of the IEEE Conference on Computer Vision and Pattern Recognition}, 2016, pp. 779--788.

\bibitem{Jocher_Ultralytics_YOLO_2023}
Glenn Jocher, Ayush Chaurasia, and Jing Qiu,
\newblock ``{Ultralytics YOLO},'' Jan. 2023,
\newblock \url{https://github.com/ultralytics/ultralytics}.

\bibitem{ren2016faster}
Shaoqing Ren, Kaiming He, Ross Girshick, and Jian Sun,
\newblock ``Faster {R-CNN}: Towards real-time object detection with region proposal networks,''
\newblock {\em IEEE Transactions on Pattern Analysis and Machine Intelligence}, vol. 39, no. 6, pp. 1137--1149, 2016.

\bibitem{he2015deep}
Kaiming He, Xiangyu Zhang, Shaoqing Ren, and Jian Sun,
\newblock ``Deep residual learning for image recognition,''
\newblock in {\em Proceedings of the IEEE Conference on Computer Vision and Pattern Recognition}, 2016, pp. 770--778.

\bibitem{howard2017mobilenets}
Andrew~G Howard, Menglong Zhu, Bo~Chen, Dmitry Kalenichenko, Weijun Wang, Tobias Weyand, Marco Andreetto, and Hartwig Adam,
\newblock ``Mobilenets: Efficient convolutional neural networks for mobile vision applications,''
\newblock {\em arXiv preprint arXiv:1704.04861}, 2017.

\bibitem{loshchilov2017decoupled}
Ilya Loshchilov and Frank Hutter,
\newblock ``Decoupled weight decay regularization,''
\newblock {\em arXiv preprint arXiv:1711.05101}, 2017.

\bibitem{rs12193152}
Luc Courtrai, Minh-Tan Pham, and Sébastien Lefèvre,
\newblock ``Small object detection in remote sensing images based on super-resolution with auxiliary generative adversarial networks,''
\newblock {\em Remote Sensing}, vol. 12, no. 19, 2020.

\bibitem{rs12152501}
Minh-Tan Pham, Luc Courtrai, Chloé Friguet, Sébastien Lefèvre, and Alexandre Baussard,
\newblock ``Yolo-fine: One-stage detector of small objects under various backgrounds in remote sensing images,''
\newblock {\em Remote Sensing}, vol. 12, no. 15, 2020.

\bibitem{alzubaidi2021review}
Laith Alzubaidi, Jinglan Zhang, Amjad~J Humaidi, Ayad Al-Dujaili, Ye~Duan, Omran Al-Shamma, Jos{\'e} Santamar{\'\i}a, Mohammed~A Fadhel, Muthana Al-Amidie, and Laith Farhan,
\newblock ``Review of deep learning: concepts, cnn architectures, challenges, applications, future directions,''
\newblock {\em Journal of big Data}, vol. 8, pp. 1--74, 2021.

\end{thebibliography}

\end{document}